\documentclass[runningheads]{llncs}
\usepackage{graphicx}
\usepackage{amsmath,amssymb} 
\usepackage{color}
\usepackage[width=122mm,left=12mm,paperwidth=146mm,height=193mm,top=12mm,paperheight=217mm]{geometry}

\usepackage{animate} 
\usepackage{xspace}
\usepackage{color}
\usepackage{multirow}
\usepackage{graphicx}
\usepackage{adjustbox}
\usepackage{subfigure}
\usepackage{paralist} 
\usepackage{enumitem} 
\usepackage{booktabs} 
\usepackage{array}
\usepackage{caption} 

\graphicspath{{figures/}}

\newcommand{\Paragraph}[1]{{\flushleft{\textbf{#1}}}} 

\definecolor{gray}{rgb}{0.35,0.35,0.35}
\definecolor{MyBlue}{rgb}{0,0.2,0.8}
\definecolor{MyRed}{rgb}{0.8,0.2,0}
\definecolor{MyGreen}{rgb}{0.0,0.4,0.1}
\definecolor{MyGray}{rgb}{0.4,0.4,0.4}

\def\red#1{\textcolor{red}{#1}}

\long\def\ignorethis#1{}

\newlength\paramargin
\newlength\figmargin
\newlength\secmargin

\newcolumntype{L}[1]{>{\raggedright\let\newline\\\arraybackslash\hspace{0pt}}m{#1}}
\newcolumntype{C}[1]{>{\centering\let\newline\\\arraybackslash\hspace{0pt}}m{#1}}
\newcolumntype{R}[1]{>{\raggedleft\let\newline\\\arraybackslash\hspace{0pt}}m{#1}}

\def\etal{et~al.\xspace}

\newcommand{\secref}[1]{Section~\ref{sec:#1}}
\newcommand{\figref}[1]{Fig.~\ref{fig:#1}}
\newcommand{\tabref}[1]{Table~\ref{tab:#1}}
\newcommand{\eqnref}[1]{\eqref{eq:#1}}

\usepackage[pagebackref=false,breaklinks=true,letterpaper=true,colorlinks,urlcolor=blue,citecolor=blue,linkcolor=blue,bookmarks=false]{hyperref}

\begin{document}
\pagestyle{headings}
\mainmatter
\def\ECCV18SubNumber{152}  

\title{Learning Blind Video Temporal Consistency} 


\authorrunning{W.-S. Lai, J.-B. Huang, O. Wang, E. Shechtman, E. Yumer, and M.-H. Yang}

\author{
    Wei-Sheng Lai$^{1}$
	\hspace{20pt}
	Jia-Bin Huang$^{2}$
	\hspace{20pt}
	Oliver Wang$^{3}$
    \hspace{20pt}
	Eli Shechtman$^{3}$
    \\
	Ersin Yumer$^{4}$
    \hspace{25pt}
	Ming-Hsuan Yang$^{1,5}$
}
\institute{
    $^1$UC Merced
    \hspace{10pt}
    $^2$Virginia Tech
    \hspace{10pt}
    $^3$Adobe Research
    \hspace{10pt}
    $^4$Argo AI
    \hspace{10pt}
    $^5$Google Cloud
}

\maketitle

\begin{figure}
    \centering
    \begin{tabular}{cc}
        \animategraphics[autoplay,loop,width=0.49\textwidth]{24}{figures/teaser/colorization/}{00011}{00059} 
        &
        \animategraphics[autoplay,loop,width=0.49\textwidth]{24}{figures/teaser/enhancement/}{00001}{00049} 
        \\
        Colorization &
        Enhancement
        \\
        \animategraphics[autoplay,loop,width=0.49\textwidth]{24}{figures/teaser/style_transfer/}{00061}{00109} 
        & 
        \animategraphics[autoplay,loop,width=0.49\textwidth]{24}{figures/teaser/intrinsic/}{00001}{00049} 
        \\
        Style transfer
        &
        Intrinsic decomposition
        \\
    \end{tabular}
    \caption{
        \textbf{Applications of the proposed method.}
        Our algorithm takes per-frame processed videos with serious temporal flickering as inputs (lower-left) and generates temporally stable videos (upper-right) while maintaining perceptual similarity to the processed frames.
		Our method is blind to the specific image processing algorithm applied to input videos and runs a high frame-rates.
		This figure contains \emph{animated videos}, which are best viewed using Adobe Acrobat.
	}
    \label{fig:teaser}
\end{figure}

\begin{abstract}
Applying image processing algorithms independently to each frame of a video often leads to undesired inconsistent results over time. 
Developing temporally consistent video-based extensions, however, requires domain knowledge for individual tasks and is unable to generalize to other applications.
In this paper, we present an efficient approach based on a deep recurrent network for enforcing temporal consistency in a video.
Our method takes the original and per-frame processed videos as inputs to produce a temporally consistent video.
Consequently, our approach is agnostic to specific image processing algorithms applied to the original video.
We train the proposed network by minimizing both short-term and long-term temporal losses as well as a perceptual loss to strike a balance between temporal coherence and perceptual similarity with the processed frames.
At test time, our model does not require computing optical flow and thus achieves real-time speed even for high-resolution videos.
We show that our single model can handle multiple and unseen tasks, including but not limited to artistic style transfer, enhancement, colorization, image-to-image translation and intrinsic image decomposition.
Extensive objective evaluation and subject study demonstrate that the proposed approach performs favorably against the state-of-the-art methods on various types of videos.
%
\end{abstract}

\section{Introduction}
\label{sec:intro}

Recent advances of deep convolutional neural networks (CNNs) have led to the development of many powerful image processing techniques including, image filtering~\cite{DJF-ECCV-2016,Xu-ICML-2015}, enhancement~\cite{DBL,LapSRN,Yan-TOG-2016}, style transfer~\cite{AdaIn,Johnson-ECCV-2016,WCT}, colorization~\cite{Iizuka-TOG-2016,Zhang-ECCV-2016}, and general image-to-image translation tasks~\cite{Pix2Pix,DRIT,CycleGAN}.
However, extending these CNN-based methods to video is non-trivial due to memory and computational constraints, and the availability of training datasets.
Applying image-based algorithms independently to each video frame typically leads to temporal flickering due to the instability of global optimization algorithms or highly non-linear deep networks.
One approach for achieving temporally coherent results is to explicitly embed flow-based temporal consistency loss in the design and training of the networks.
However, such an approach suffers from two drawbacks.
First, it requires domain knowledge to re-design the algorithm~\cite{Aydin-TOG-2014,Huang-TOG-2014}, re-train a deep model~\cite{Gupta-ICCV-2017,Huang-CVPR-2017}, and video datasets for training.
Second, due to the dependency of flow computation at test time, these approaches tend to be slow. 
%

Bonneel~\etal~\cite{Bonneel-TOG-2015} propose a general approach to achieve temporal coherent results that is \emph{blind} to specific image processing algorithms.
The method takes the original video and the per-frame processed video as inputs and solves a gradient-domain optimization problem to minimize the temporal warping error between consecutive frames.
Although the results of Bonneel~\etal~\cite{Bonneel-TOG-2015} are temporally stable, their algorithm highly depends on the quality of dense correspondence (e.g., optical flow or PatchMatch~\cite{PatchMatch}) and may fail when a severe occlusion occurs.
Yao~\etal~\cite{Yao-MM-2017} further extend the method of Bonneel~\etal~\cite{Bonneel-TOG-2015} to account for occlusion by selecting a set of key-frames.
However, the computational cost increases linearly with the number of key-frames, and thus their approach cannot be efficiently applied to long video sequences.
Furthermore, both approaches assume that the gradients of the original video are similar to the gradients of the processed video, which restricts them from handling tasks that may generate new contents (e.g., stylization).

In this work, we formulate the problem of video temporal consistency as a learning task.
We propose to learn a deep recurrent network that takes the input and processed videos and generates temporally stable output videos.
We minimize the short-term and long-term temporal losses between output frames and impose a perceptual loss from the pre-trained VGG network~\cite{VGG} to maintain the perceptual similarity between the output and processed frames.
In addition, we embed a convolutional LSTM (ConvLSTM)~\cite{convlstm} layer to capture the spatial-temporal correlation of natural videos.
Our network processes video frames sequentially and can be applied to videos with arbitrary lengths.
Furthermore, our model does not require computing optical flow at \emph{test} time and thus can process videos at real-time rates ($400+$ FPS on $1280 \times 720$ videos).

As existing video datasets typically contain low-quality frames, we collect a high-quality video dataset with 80 videos for training and 20 videos for evaluation.
We train our model on a wide range of applications, including colorization, image enhancement, and artistic style transfer, and demonstrate that a \emph{single} trained model generalizes well to \emph{unseen} applications (e.g., intrinsic image decomposition, image-to-image translation).
We evaluate the quality of the output videos using temporal warping error and a learned perceptual metric~\cite{Zhang-CVPR-2018}. 
We show that the proposed method strikes a good balance between maintaining the temporal stability and perceptual similarity.
Furthermore, we conduct a user study to evaluate the subjective preference between the proposed method and state-of-the-art approaches.

We make the following contributions in this work:
\begin{enumerate}
\item We present an efficient solution to remove temporal flickering in videos via learning a deep network with a ConvLSTM module.
Our method does not require pre-computed optical flow or frame correspondences at \emph{test} time and thus can process videos in real-time.

\item We propose to minimize the short-term and long-term temporal loss for improving the temporal stability and adopt a perceptual loss to maintain the perceptual similarity. 

\item We provide a \emph{single} model for handling multiple applications, including but not limited to colorization, enhancement, artistic style transfer, image-to-image translation and intrinsic image decomposition.
Extensive subject and objective evaluations demonstrate that the proposed algorithm performs favorably against existing approaches on various types of videos.
	
\end{enumerate}

\section{Related Work}
\label{sec:related}
We address the temporal consistency problem on a wide range of applications, including automatic white balancing~\cite{Hsu-TOG-2008}, harmonization~\cite{Bonneel-2015}, dehazing~\cite{He-PAMI-2011}, image enhancement~\cite{DBL}, style transfer~\cite{AdaIn,Johnson-ECCV-2016,WCT}, colorization~\cite{Iizuka-TOG-2016,Zhang-ECCV-2016}, image-to-image translation~\cite{Pix2Pix,CycleGAN}, and intrinsic image decomposition~\cite{Bell-TOG-2014}.
A complete review of these applications is beyond the scope of this paper.
In the following, we discuss task-specific and task-independent approaches that enforce temporal consistency on videos.

\Paragraph{Task-specific approaches.}
A common solution to embed the temporal consistency constraint is to use optical flow to propagate information between frames, e.g., colorization~\cite{Levin-TOG-2004} and intrinsic decomposition~\cite{Ye-TOG-2014}. 
However, estimating optical flow is computationally expensive and thus is impractical to apply on high-resolution and long sequences.
Temporal filtering is an efficient approach to extend image-based algorithms to videos, e.g., tone-mapping~\cite{Aydin-TOG-2014}, color transfer~\cite{Bonneel-TOG-2013}, and visual saliency~\cite{Lang-TOG-2012} to generate temporally consistent results.
Nevertheless, these approaches assume a specific filter formulation and cannot be generalized to other applications.

Recently, several approaches have been proposed to improve the temporal stability of CNN-based image style transfer.
Huang~\etal~\cite{Huang-CVPR-2017} and Gupta~\etal~\cite{Gupta-ICCV-2017} train feed-forward networks by jointly minimizing content, style and temporal warping losses.
These methods, however, are limited to the specific styles used during training.
Chen~\etal~\cite{Chen-ICCV-2017} learn flow and mask networks to adaptively blend the intermediate features of the pre-trained style network.
While the architecture design is independent of the style network, it requires the access to intermediate features and cannot be applied to non-differentiable tasks.
In contrast, the proposed model is entirely blind to specific algorithms applied to the input frames and thus is applicable to optimization-based techniques, CNN-based algorithms, and combinations of Photoshop filters.

\begin{table}[t]
	\centering
	\scriptsize
	\caption{
		\textbf{Comparison of blind temporal consistency methods}.
		Both the methods of Bonneel~\etal~\cite{Bonneel-TOG-2015} and Yao~\etal~\cite{Yao-MM-2017} require dense correspondences from optical flow or PatchMatch~\cite{PatchMatch}, while the proposed method does not explicitly rely on these correspondences at test time. 
		%
		The algorithm of Yao~\etal~\cite{Yao-MM-2017} involves a key-frame selection from the entire video and thus cannot generate output in an online manner.
	}
	\begin{tabular}{c|C{2.5cm}C{2.5cm}C{2.5cm}}
		\toprule
		& Bonneel~\etal~\cite{Bonneel-TOG-2015} & Yao~\etal~\cite{Yao-MM-2017} & Ours
		\\
		\midrule
		Content constraint &
		gradient & local affine & perceptual loss \\
		Short-term temporal constraint &
		\checkmark & - & \checkmark \\
		Long-term temporal constraint &
		- & \checkmark & \checkmark \\
		Require dense correspondences & 
		\multirow{2}{*}{\checkmark} & \multirow{2}{*}{\checkmark} & \multirow{2}{*}{-} \\
		(at test time) & & & \\
		Online processing &
		\checkmark & - & \checkmark \\
		\bottomrule
	\end{tabular}
	\label{tab:blind_comparison}
\end{table}

\Paragraph{Task-independent approaches.}
Several methods have been proposed to improve temporal consistency for multiple tasks.
Lang~\etal~\cite{Lang-TOG-2012} approximate global optimization of a class of energy formulation (e.g., colorization, optical flow estimation) via temporal edge-aware filtering.
In~\cite{Dong-CVPR-2015}, Dong~\etal propose a segmentation-based algorithm and assume that the image transformation is spatially and temporally consistent.
More general approaches assume gradient similarity~\cite{Bonneel-TOG-2015} or local affine transformation~\cite{Yao-MM-2017} between the input and the processed frames.
These methods, however, cannot handle more complicated tasks (e.g., artistic style transfer).
In contrast, we use the VGG perceptual loss~\cite{Johnson-ECCV-2016} to impose high-level perceptual similarity between the output and processed frames.
We list the feature-by-feature comparisons between Bonneel~\etal~\cite{Bonneel-TOG-2015}, Yao~\etal~\cite{Yao-MM-2017} and the proposed method in~\tabref{blind_comparison}.

\section{Learning Temporal Consistency}
\label{sec:method}
%
In this section, we describe the proposed recurrent network and the design of the loss functions for enforcing temporal consistency on videos.

\subsection{Recurrent network}
\label{sec:RNN}
\begin{figure}[t]
	\centering
	\includegraphics[width=0.9\textwidth]{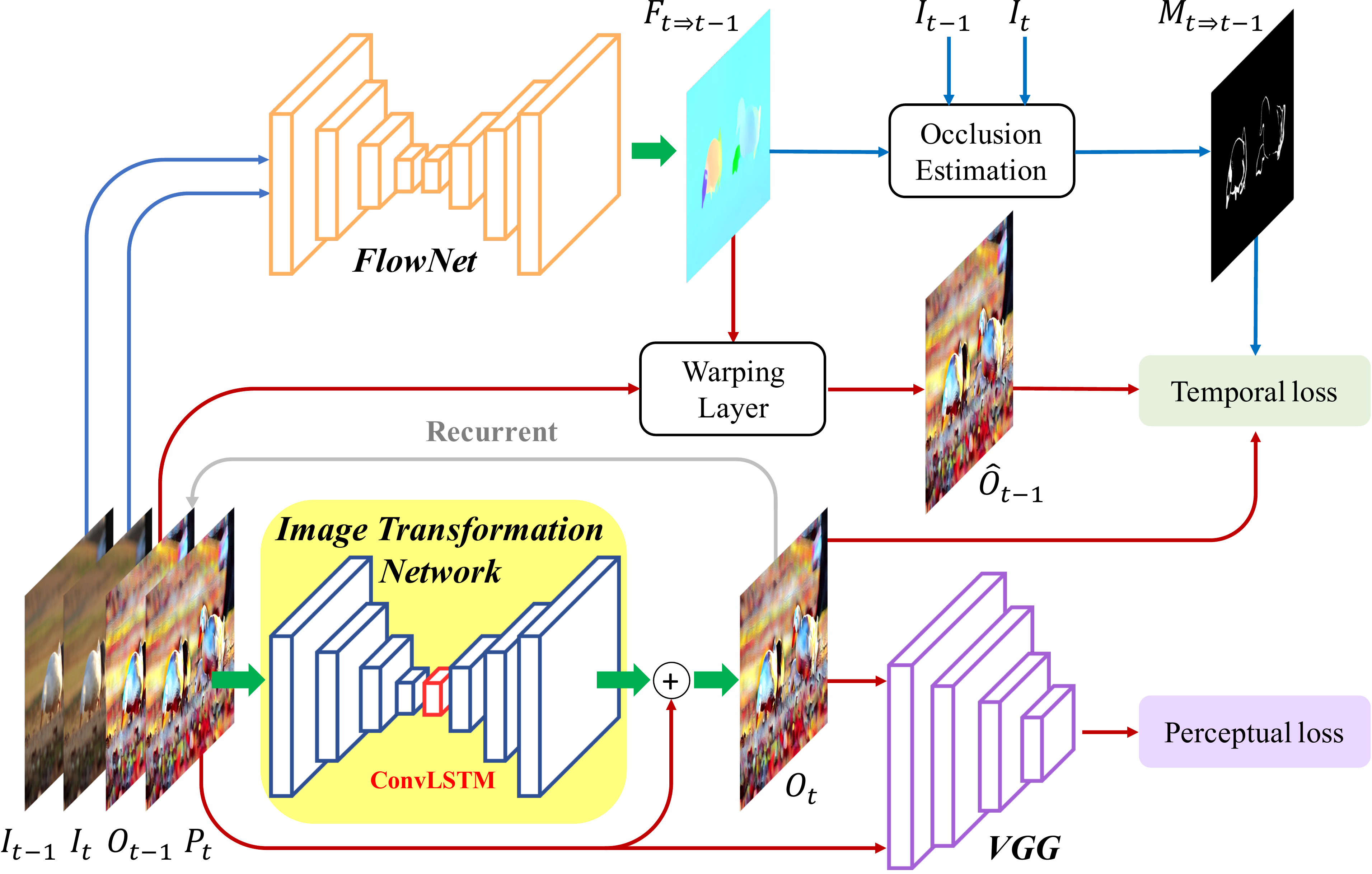}
	\caption{
		\textbf{Overview of the proposed method.}
		We train an image transformation network that takes $I_{t-1}, I_t, O_{t-1}$ and processed frame $P_t$ as inputs and generates the output frame $O_t$ which is temporally consistent with the output frame at the previous time step $O_{t-1}$.
		The output $O_t$ at the current time step then becomes the input at the next time step.
		We train the image transformation network with the VGG perceptual loss and the short-term and long-term temporal losses.
	}
	\label{fig:training}
\end{figure}

\figref{training} shows an overview of the proposed recurrent network.
Our model takes as input the original (unprocessed) video $\{I_t | t=1 \cdots T\}$ and per-frame processed videos $\{P_t | t=1 \cdots T\}$, and produces temporally consistent output videos $\{O_t | t=1 \cdots T\}$.
In order to efficiently process videos with arbitrary length, we develop an image transformation network as a \emph{recurrent convolutional network} to generate output frames in an online manner (i.e., sequentially from $t = 1$ to $T$).
Specifically, we set the first output frame $O_1 = P_1$.
In each time step, the network learns to generate an output frame $O_t$ that is temporally consistent with respect to $O_{t-1}$.
The current output frame is then fed as the input at the next time step.
To capture the spatial-temporal correlation of videos, we integrate a ConvLSTM layer~\cite{convlstm} into our image transformation network.
We discuss the detailed design of our image transformation network in~\secref{TransformNet}.

\subsection{Loss functions}
\label{sec:loss}
Our goal is to reduce the temporal inconsistency in the output video while maintaining the perceptual similarity with the processed frames.
Therefore, we propose to train our model with (1) a perceptual content loss between the output frame and the processed frame and (2) short-term and long-term temporal losses between output frames.

\Paragraph{Content perceptual loss.}
We compute the similarity between $O_t$ and $P_t$ using the perceptual loss from a pre-trained VGG classification network~\cite{VGG}, which is commonly adopted in several applications (e.g., style transfer~\cite{Johnson-ECCV-2016}, super-resolution~\cite{SRGAN}, and image inpainting~\cite{ContextEncoder}) and has been shown to correspond well to human perception~\cite{Zhang-CVPR-2018}.
The perceptual loss is defined as:
\begin{equation}
	\mathcal{L}_p = \sum_{t=2}^T \sum_{i=1}^N \sum_{l} \left\| \phi_l(O_t^{(i)}) - \phi_l(P_t^{(i)}) \right\|_1, 
	\label{eq:perceptual_loss}
\end{equation}
where $O_t^{(i)}$ represents a vector $\in R^3$ with RGB pixel values of the output $O$ at time $t$, $N$ is the total number of pixels in a frame, and $\phi_l(\cdot)$ denotes the feature activation at the $l$-th layer of the VGG-19 network $\phi$.
We choose the 4-th layer (i.e., \texttt{relu4-3}) to compute the perceptual loss.

\begin{figure}[t]
	\centering
	\includegraphics[width=0.85\textwidth]{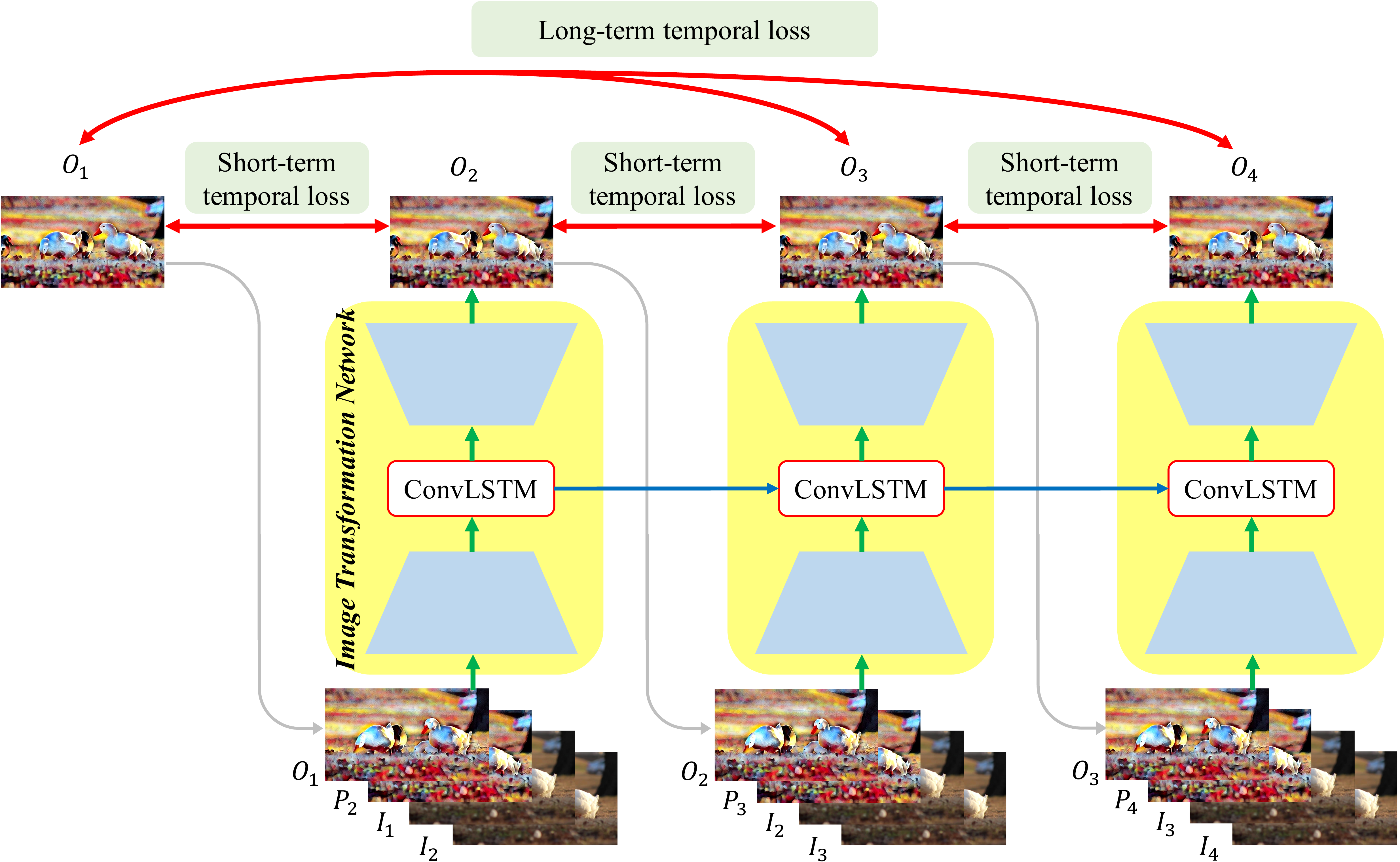}
	\caption{
		\textbf{Temporal losses.}
		We adopt the short-term temporal loss on neighbor frames and long-term temporal loss between the first and all the output frames.
	}
	\label{fig:LSTM}
\end{figure}

\Paragraph{Short-term temporal loss.}
We formulate the temporal loss as the warping error between the output frames:
\begin{align}
	\mathcal{L}_{st} = \sum_{t=2}^T \sum_{i=1}^N M_{t \Rightarrow t-1}^{(i)}  \left\| O_t^{(i)} - \hat{O}_{t-1}^{(i)} \right\|_1,
\label{eq:short_term_loss}
\end{align}
where $\hat{O}_{t-1}$ is the frame $O_{t-1}$ warped by the optical flow $F_{t \Rightarrow t-1}$, and $M_{t \Rightarrow t-1} = \exp(-\alpha \| I_t - \hat{I}_{t-1} \|_2^2)$ is the visibility mask calculated from the warping error between input frames $I_t$ and warped input frame $\hat{I}_{t-1}$.
The optical flow $F_{t \Rightarrow t-1}$ is the backward flow between $I_{t-1}$ and $I_t$.
We use the FlowNet2~\cite{FlowNet2} to efficiently compute flow on-the-fly during training.
We use the bilinear sampling layer~\cite{STN} to warp frames and empirically set $\alpha = 50$ (with pixel range between $[0, 1]$).

\Paragraph{Long-term temporal loss.}
While the short-term temporal loss~\eqnref{short_term_loss} enforces the temporal consistency between consecutive frames, there is no guarantee for long-term (e.g., more than 5 frames) coherence.
A straightforward method to enforce long-term temporal consistency is to apply the temporal loss on \emph{all} pairs of output frames.
However, such a strategy requires significant computational costs (e.g., optical flow estimation) during training.
Furthermore, computing temporal loss between two intermediate outputs is not meaningful before the network converges.

Instead, we propose to impose long-term temporal losses between the \emph{first} output frame and all of the output frames:
\begin{align}
	\mathcal{L}_{lt} = \sum_{t=2}^T \sum_{i=1}^N M_{t \Rightarrow 1}^{(i)} \left\| O_t^{(i)} - \hat{O}_{1}^{(i)} \right\|_1.
\label{eq:long_term_loss}
\end{align}
We illustrate an unrolled version of our recurrent network as well as the short-term and long-term losses in~\figref{LSTM}.
%
During the training, we enforce the long-term temporal coherence over a maximum of 10 frames ($T = 10$).

\Paragraph{Overall loss.}
The overall loss function for training our image transformation network is defined as:
\begin{align}
	\mathcal{L} = \lambda_p \mathcal{L}_p + \lambda_{st} \mathcal{L}_{st} + \lambda_{lt} \mathcal{L}_{lt}, 
\label{eq:overall_loss}
\end{align}
where $\lambda_p$, $\lambda_{st}$ and $\lambda_{lt}$ are the weights for the content perceptual loss, short-term and long-term losses, respectively.
%

\subsection{Image transformation network}
\label{sec:TransformNet}
The input of our image transformation network is the concatenation of the currently processed frame $P_t$, previous output frame $O_{t-1}$ as well as the current and previous unprocessed frames $I_t, I_{t-1}$.
As the output frame typically looks similar to the currently processed frame, we train the network to predict the \emph{residuals} instead of actual pixel values, i.e., $O_t = P_t + \mathcal{F}(P_t)$, where $\mathcal{F}$ denotes the image transformation network.
Our image transformation network consists of two strided convolutional layers, $B$ residual blocks, one ConvLSTM layer, and two transposed convolutional layers.

\begin{figure}[t]
	\footnotesize
	\centering
	\includegraphics[width=0.6\textwidth]{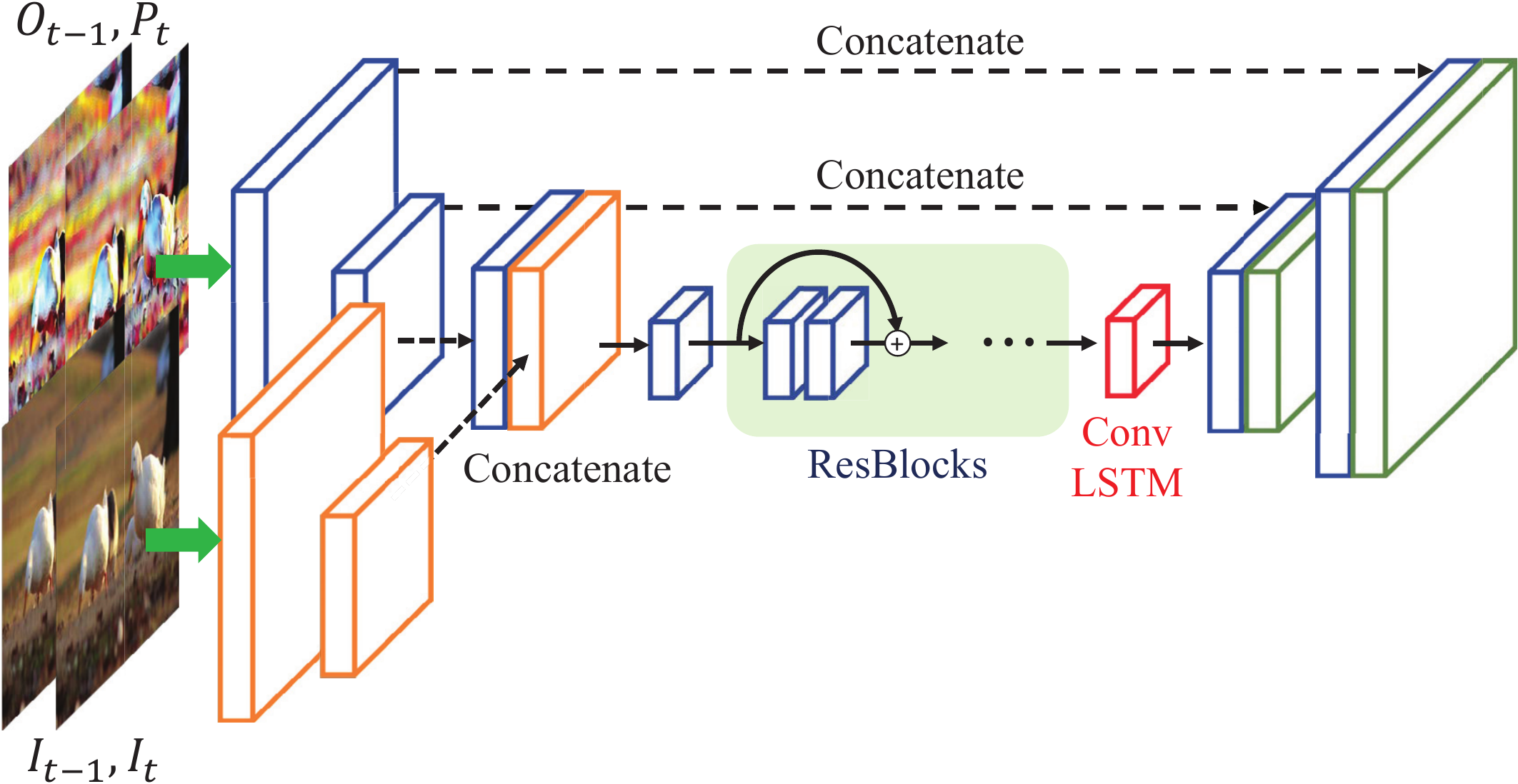}    
	\caption{
		\textbf{Architecture of our image transformation network.}
		We split the input into two streams to avoid transferring low-level information from the input frames to output.
	}
	\label{fig:TransformNet}
\end{figure}

We add skip connections from the encoder to the decoder to improve the reconstruction quality.
However, for some applications, the processed frames may have a dramatically different appearance than the input frames (e.g., style transfer or intrinsic image decomposition).
We observe that the skip connections may transfer low-level information (e.g., color) to the output frames and produce visual artifacts.
Therefore, we divide the input into two streams: one for the processed frames $P_t$ and $O_{t-1}$, and the other stream for input frames $I_t$ and $I_{t-1}$.
As illustrated in~\figref{TransformNet}, the skip connections only add skip connections from the processed frames to avoid transferring the low-level information from the input frames.
We provide all the implementation details in the supplementary material.

\section{Experimental Results}
\label{sec:experiment}
In this section, we first describe the employed datasets for training and testing, followed by the applications of the proposed method and the metrics for evaluating the temporal stability and perceptual similarity.
We then analyze the effect of each loss term in balancing the temporal coherence and perceptual similarity, conduct quantitative and subjective comparisons with existing approaches, and finally discuss the limitations of our method.
The source code and datasets are publicly available at~\url{http://vllab.ucmerced.edu/wlai24/video_consistency}.

\subsection{Datasets}
\label{sec:dataset}
We use the \textsc{DAVIS}-2017 dataset~\cite{DAVIS}, which is designed for video segmentation and contains a variety of moving objects and motion types.
The \textsc{DAVIS} dataset has 60 videos for training and 30 videos for validation.
However, the lengths of the videos in the \textsc{DAVIS} dataset are usually short (less than 3 seconds) with 4,209 training frames in total.
Therefore, we collect additional 100 high-quality videos from Videvo.net~\cite{videvo}, where 80 videos are used for training and 20 videos for testing.
We scale the height of video frames to 480 and keep the aspect ratio.
We use both the \textsc{DAVIS} and \textsc{Videvo} training sets, which contains a total of 25,735 frames, to train our network.

\subsection{Applications}
\label{sec:application}
As we do not make any assumptions on the underlying image-based algorithms, our method is applicable for handling a wide variety of applications.

\Paragraph{Artistic style transfer.}
The tasks of image style transfer have been shown to be sensitive to minor changes in content images due to the non-convexity of the Gram matrix matching objective~\cite{Gupta-ICCV-2017}.
We apply our method to the results from the state-of-the-art style transfer approaches~\cite{Johnson-ECCV-2016,WCT}.

\Paragraph{Colorization.}
Single image colorization aims to hallucinate plausible colors from a given grayscale input image.
Recent algorithms~\cite{Iizuka-TOG-2016,Zhang-ECCV-2016} learn deep CNNs from millions of natural images.
When applying colorization methods to a video frame-by-frame, those approaches typically produce low-frequency flickering.

\Paragraph{Image enhancement.} 
Gharbi~\etal~\cite{DBL} train deep networks to learn the user-created action scripts of Adobe Photoshop for enhancing images.
Their models produce high-frequency flickering on most of the videos.

\Paragraph{Intrinsic image decomposition.} 
Intrinsic image decomposition aims to decompose an image into a reflectance and a shading layer.
The problem is highly ill-posed due to the scale ambiguity.
We apply the approach of Bell~\etal~\cite{Bell-TOG-2014} to our test videos. 
As expected, the image-based algorithm produces serious temporal flickering artifacts when applied to each frame in the video independently.

\Paragraph{Image-to-image translation.} 
In recent years, the image-to-image translation tasks attract considerable attention due to the success of the Generative Adversarial Networks (GAN)~\cite{GAN}.
The CycleGAN model~\cite{CycleGAN} aims to learn mappings from one image domain to another domain without using paired training data.
When the transformations generate a new texture on images (e.g., photo $\rightarrow$ painting, horse$ \rightarrow$ zebra) or the mapping contains multiple plausible solutions (e.g., gray $\rightarrow$ RGB), the resulting videos inevitably suffer from temporal flickering artifacts.

The above algorithms are general and can be applied to any type of videos. When applied, they produce temporal flickering artifacts on most videos in our test sets.
We use the WCT~\cite{WCT} style transfer algorithm with three style images, one of the enhancement models of Gharbi~\etal~\cite{DBL}, the colorization method of Zhang~\etal~\cite{Zhang-ECCV-2016} and the shading layer of Bell~\etal~\cite{Bell-TOG-2014} as our training tasks, with the rest of the tasks being used for testing purposes.
We demonstrate that the proposed method learns a \emph{single} model for multiple applications and also generalizes to \emph{unseen} tasks.

\subsection{Evaluation metrics}
Our goal is to generate a temporally smooth video while maintaining the perceptual similarity with the per-frame processed video.
We use the following metrics to measure the temporal stability and perceptual similarity on the output videos.

\Paragraph{Temporal stability.}
We measure the temporal stability of a video based on the flow warping error between two frames:
\begin{equation}
	E_{\text{warp}}(V_t, V_{t+1}) = \frac{1}{\sum_{i=1}^N M_t^{(i)}} \sum_{i=1}^N M_t^{(i)} \| V_t^{(i)} - \hat{V}_{t+1}^{(i)} \|_2^2, 
	\label{eq:frame_warp_error}
\end{equation}
where $\hat{V}_{t+1}$ is the warped frame $V_{t+1}$ and $M_t \in \{0, 1\}$ is a non-occlusion mask indicating non-occluded regions.
We use the occlusion detection method in~\cite{Ruder-2016} to estimate the mask $M_t$.
The warping error of a video is calculated as:
\begin{equation}
	E_{\text{warp}}(V) = \frac{1}{T-1} \sum_{t=1}^{T-1} E_{\text{warp}}(V_t, V_{t+1}), 
	\label{eq:temporal_warp_error}
\end{equation}
which is the average warping error over the entire sequence.

\Paragraph{Perceptual similarity.}
Recently, the features of the pre-trained VGG network~\cite{VGG} have been shown effective as a training loss to generate realistic images in several vision tasks~\cite{Chen-ICCV-2017b,SRGAN,ContextEncoder}.
Zhang~\etal~\cite{Zhang-CVPR-2018} further propose a perceptual metric by calibrating the deep features of ImageNet classification networks.
We adopt the calibrated model of the SqueezeNet~\cite{Squeezenet} (denote as $\mathcal{G}$) to measure the perceptual distance of the processed video $P$ and output video $O$:
\begin{equation}
	D_{\text{perceptual}}(P, O) = \frac{1}{T-1} \sum_{t=2}^{T} \mathcal{G}(O_t, P_t).
	\label{eq:perceptual_distance}
\end{equation}
We note that the first frame is fixed as a reference in both Bonneel~\cite{Bonneel-TOG-2015} and our algorithm.
Therefore, we exclude the first frame from computing the perceptual distance in~\eqnref{perceptual_distance}.

\begin{figure}[t]
    \centering
    \scriptsize
    \begin{minipage}[b]{0.47\textwidth}
        \centering
        \begin{tabular}{C{0.8cm}C{0.8cm}C{0.9cm}|C{1.4cm}C{1.4cm}}
            \toprule
            $\lambda_t$ & $\lambda_p$ & $r= \frac{\lambda_t}{\lambda_p}$ &
            $E_{\text{warp}}$ & $D_{\text{perceptual}}$
            \\
            \midrule
            10    &  0.01  &  1000  &  0.0279 & 0.1744 \\
            10    &  0.1   &  100   &  0.0265 & 0.1354 \\
            10    &  1     &  10    &  0.0615 & 0.0071 \\
            10    &  10    &  1     &  0.0621 & 0.0072 \\
            100   &  1     &  100   &  0.0277 & 0.1324 \\
            100   &  10    &  10    &  \red{0.0442} & \red{0.0170} \\
            100   &  100   &  1     &  0.0621 & 0.0072 \\
            1000  &  1     &  1000  &  0.0262 & 0.1848 \\
            1000  &  10    &  100   &  0.0275 & 0.1341 \\
            1000  &  100   &  10    &  \red{0.0453} & \red{0.0158} \\
            1000  &  1000  &  1     &  0.0621 & 0.0072 \\
            \bottomrule
        \end{tabular}
        \label{tab:trade_off}
    \end{minipage}
    \hfill
    \begin{minipage}[b]{0.47\textwidth}
        \centering
        \includegraphics[width=0.9\textwidth]{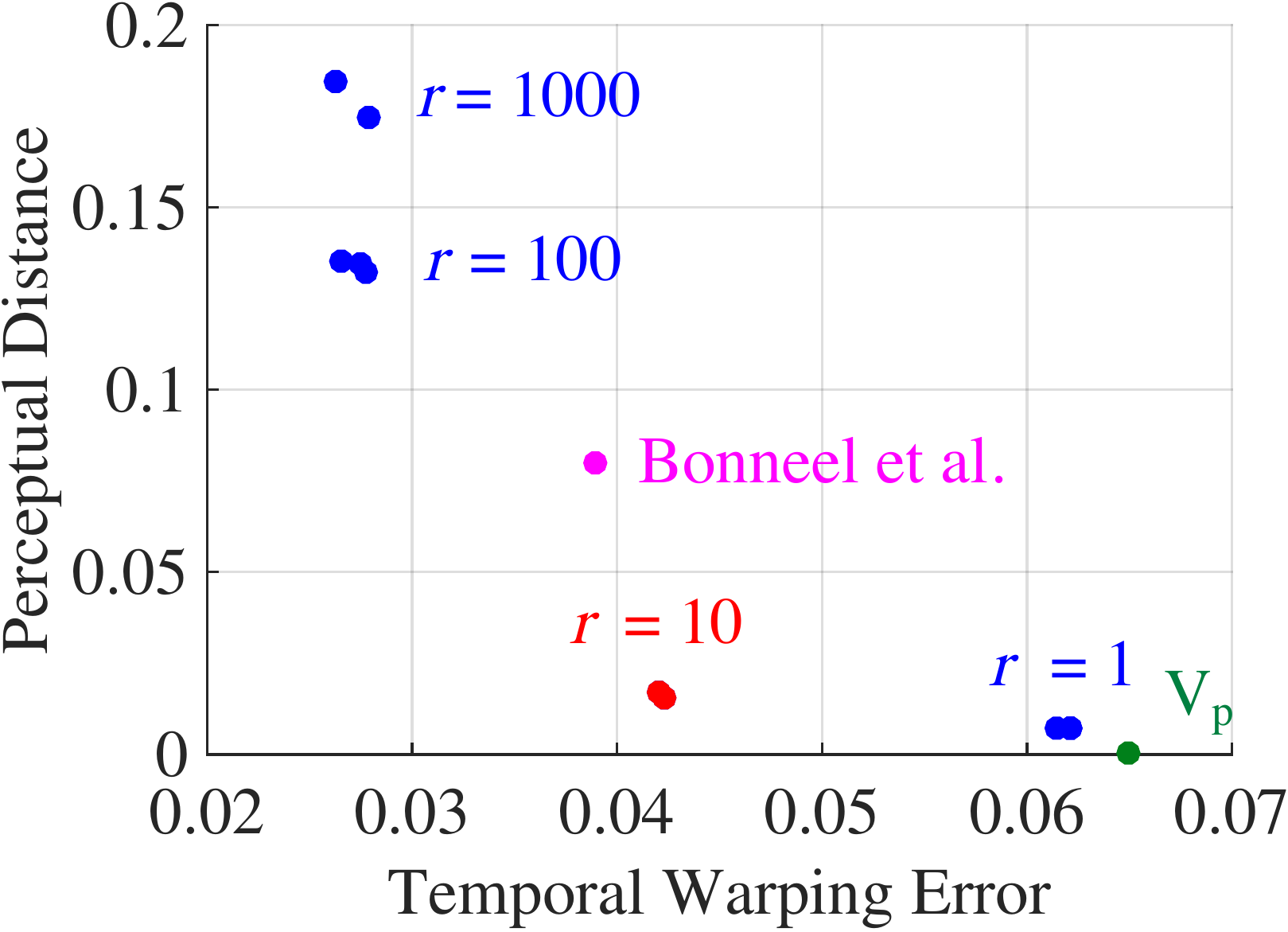} 
    \end{minipage}
    \caption{
        \textbf{Analysis of parameters}.
        (Left) When $\lambda_t$ is large enough, choosing $r = 10$ (shown in red) achieves a good balance between reducing temporal warping error as well as perceptual distance.
        (Right) The trade off between perceptual similarity and temporal warping with different ratios $r$, as compared to Bonneel et al.~\protect\cite{Bonneel-TOG-2015}, and the original processed video, $V_p$.
    }
    \label{fig:trade_off}
\end{figure}

\subsection{Analysis and discussions}
\label{sec:ablation}
An extremely blurred video may have high temporal stability but with low perceptual similarity; in contrast, the processed video itself has perfect perceptual similarity but is temporally unstable.
Due to the trade-off between the temporal stability and perceptual similarity, it is important to balance these two properties and produce visually pleasing results.

To understand the relationship between the temporal and content losses, we train models with the several combinations of $\lambda_p$ and $\lambda_t$ ($= \lambda_{st} = \lambda_{lt}$).
We use one of the styles (i.e., udnie) from the fast neural style transfer method~\cite{Johnson-ECCV-2016} for evaluation.
We show the quantitative evaluation on the \textsc{DAVIS} test set in~\figref{trade_off}.
We observe that the ratio $r = \lambda_t / \lambda_p$ plays an important role in balancing the temporal stability and perceptual similarity.
When the ratio $r < 10$, the perceptual loss dominates the optimization of the network, and the temporal flickering remains in the output videos.
When the ratio $r > 10$, the output videos become overly blurred and therefore have a large perceptual distance to the processed videos.
When $\lambda_{t}$ is sufficiently large (i.e., $\lambda_t \geq 100$), the setting $r = 10$ strikes a good balance to reduce temporal flickering while maintaining small perceptual distance.
Our results find similar observation on other applications as well.

\begin{table}[t]
	\centering
	\scriptsize
	\caption{
		\textbf{Quantitative evaluation on temporal warping error}.
		The ``Trained'' column indicates the applications used for training our model.
		Our method achieves a similarly reduced temporal warping error as Bonneel~\etal~\cite{Bonneel-TOG-2015}, which is significantly less than the original processed video ($V_p$). 
	}
	\begin{tabular}{l|c|C{1cm}C{1cm}C{1cm}|C{1cm}C{1cm}C{1cm}}
		\toprule
		\multirow{2}{*}{Task} &
		\multirow{2}{*}{Trained} &
		\multicolumn{3}{c|}{\textsc{Davis}} &
		\multicolumn{3}{c}{\textsc{Videvo}} \\
		& & $V_p$
		& \cite{Bonneel-TOG-2015} 
		& Ours 
		& $V_p$
		& \cite{Bonneel-TOG-2015} 
		& Ours 
		\\
		\midrule
		WCT~\cite{WCT}/antimono & \checkmark &
		0.054 & \textbf{0.031} & 0.035 & 0.025 & 0.014 & \textbf{0.013}
		\\
		WCT~\cite{WCT}/asheville & &
		0.088 & \textbf{0.047} & 0.055 & 0.045 & 0.025 & \textbf{0.023}
		\\
		WCT~\cite{WCT}/candy & \checkmark &
		0.069 & \textbf{0.037} & 0.045 & 0.034 & \textbf{0.018} & \textbf{0.018}
		\\
		WCT~\cite{WCT}/feathers & &
		0.052 & \textbf{0.029} & \textbf{0.029} & 0.027 & 0.016 & \textbf{0.012}
		\\
		WCT~\cite{WCT}/sketch & \checkmark &
		0.046 & 0.028 & \textbf{0.023} & 0.022 & 0.015 & \textbf{0.009}
		\\
		WCT~\cite{WCT}/wave & &
		0.049 & 0.030 & \textbf{0.027} & 0.026 & 0.015 & \textbf{0.011}
		\\
		Fast-neural-style~\cite{Johnson-ECCV-2016}/princess &  &
		0.073 & 0.048 & \textbf{0.047} & 0.039 & 0.023 & \textbf{0.021}
		\\
		Fast-neural-style~\cite{Johnson-ECCV-2016}/udnie &  &
		0.065 & \textbf{0.039} & 0.042 & 0.028 & 0.017 & \textbf{0.015}
		\\
		DBL~\cite{DBL}/expertA & \checkmark &
		0.039 & 0.035 & \textbf{0.028} & 0.018 & 0.016 & \textbf{0.010}
		\\
		DBL~\cite{DBL}/expertB & &
		0.034 & 0.031 & \textbf{0.025} & 0.015 & 0.014 & \textbf{0.008}
		\\
		Intrinsic~\cite{Bell-TOG-2014}/reflectance &  &
		0.024 & 0.020 & \textbf{0.015} & 0.012 & 0.008 & \textbf{0.005}
		\\
		Intrinsic~\cite{Bell-TOG-2014}/shading & \checkmark &
		0.016 & 0.012 & \textbf{0.009} & 0.008 & 0.006 & \textbf{0.003}
		\\
		CycleGAN~\cite{CycleGAN}/photo2ukiyoe &  &
		0.037 & 0.030 & \textbf{0.026} & 0.019 & 0.016 & \textbf{0.010}
		\\
		CycleGAN~\cite{CycleGAN}/photo2vangogh &  &
		0.040 & 0.032 & \textbf{0.029} & 0.021 & 0.017 & \textbf{0.013}
		\\
		Colorization~\cite{Zhang-ECCV-2016} & \checkmark &
		0.030 & 0.028 & \textbf{0.024} & 0.012 & 0.011 & \textbf{0.008}
		\\
		Colorization~\cite{Iizuka-TOG-2016} & &
		0.030 & 0.028 & \textbf{0.023} & 0.012 & 0.011 & \textbf{0.008}
		\\
		\midrule
		Average & &
		0.047 & 0.032 & \textbf{0.030} & 0.023 & 0.015 & \textbf{0.012}
		\\
		\bottomrule
	\end{tabular}
	\label{tab:warp_error}
\end{table}

\begin{table}
	\centering
	\scriptsize
	\caption{
		\textbf{Quantitative evaluation on perceptual distance}.
		Our method has lower perceptual distance than Bonneel~\etal~\cite{Bonneel-TOG-2015}.
	}
	\begin{tabular}{l|c|C{1cm}C{1cm}|C{1cm}C{1cm}}
		\toprule
		\multirow{2}{*}{Task} &
		\multirow{2}{*}{Trained} &
		\multicolumn{2}{c|}{\textsc{Davis}} &
		\multicolumn{2}{c}{\textsc{Videvo}} \\
		& 
		& \cite{Bonneel-TOG-2015} 
		& Ours 
		& \cite{Bonneel-TOG-2015} 
		& Ours 
		\\
		\midrule
		WCT~\cite{WCT}/antimono & \checkmark &
		0.098 & \textbf{0.019} & 0.106 & \textbf{0.016}
		\\
		WCT~\cite{WCT}/asheville &  &
		0.090 & \textbf{0.019} & 0.098 & \textbf{0.015}
		\\
		WCT~\cite{WCT}/candy & \checkmark &
		0.133 & \textbf{0.023} & 0.139 & \textbf{0.018}
		\\
		WCT~\cite{WCT}/feathers & &
		0.093 & \textbf{0.016} & 0.100 & \textbf{0.011}
		\\
		WCT~\cite{WCT}/sketch & \checkmark &
		0.042 & \textbf{0.021} & 0.046 & \textbf{0.014}
		\\
		WCT~\cite{WCT}/wave & &
		0.065 & \textbf{0.015} & 0.072 & \textbf{0.013}
		\\
		Fast-neural-style~\cite{Johnson-ECCV-2016}/princess &  &
		0.143 & \textbf{0.029} & 0.165 & \textbf{0.018}
		\\
		Fast-neural-style~\cite{Johnson-ECCV-2016}/udnie &  &
		0.070 & \textbf{0.017} & 0.076 & \textbf{0.014}
		\\
		DBL~\cite{DBL}/expertA & \checkmark &
		0.026 & \textbf{0.011} & 0.033 & \textbf{0.007}
		\\
		DBL~\cite{DBL}/expertB & &
		0.023 & \textbf{0.011} & 0.030 & \textbf{0.007}
		\\
		Intrinsic~\cite{Bell-TOG-2014}/reflectance &  &
		0.044 & \textbf{0.013} & 0.056 & \textbf{0.008}
		\\
		Intrinsic~\cite{Bell-TOG-2014}/shading & \checkmark &
		0.029 & \textbf{0.017} & 0.032 & \textbf{0.009}
		\\
		CycleGAN~\cite{CycleGAN}/photo2ukiyoe &  &
		0.042 & \textbf{0.012} & 0.054 & \textbf{0.007}
		\\
		CycleGAN~\cite{CycleGAN}/photo2vangogh &  &
		0.067 & \textbf{0.016} & 0.079 & \textbf{0.011}
		\\
		Colorization~\cite{Zhang-ECCV-2016} & \checkmark &
		0.062 & \textbf{0.013} & 0.055 & \textbf{0.009}
		\\
		Colorization~\cite{Iizuka-TOG-2016} & &
		0.033 & \textbf{0.011} & 0.034 & \textbf{0.008}
		\\
		\midrule
		Average & &
		0.088 & \textbf{0.017} & 0.073 & \textbf{0.012}
		\\
		\bottomrule
	\end{tabular}
	\label{tab:perceptual}
\end{table}

\begin{figure}[t]
	\centering
	\scriptsize
	\begin{tabular}{cccc}
	    \includegraphics[width=0.24\textwidth]{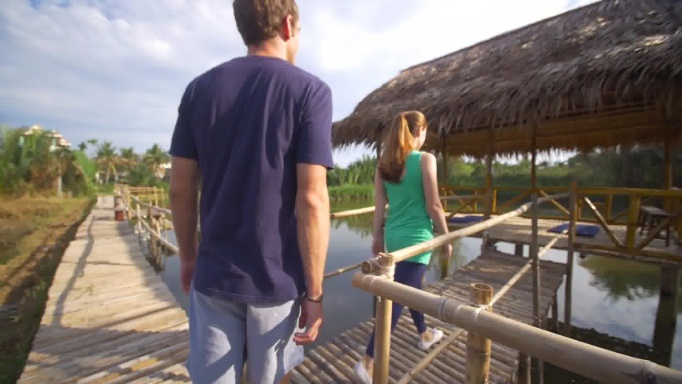}
	    &
	    \includegraphics[width=0.24\textwidth]{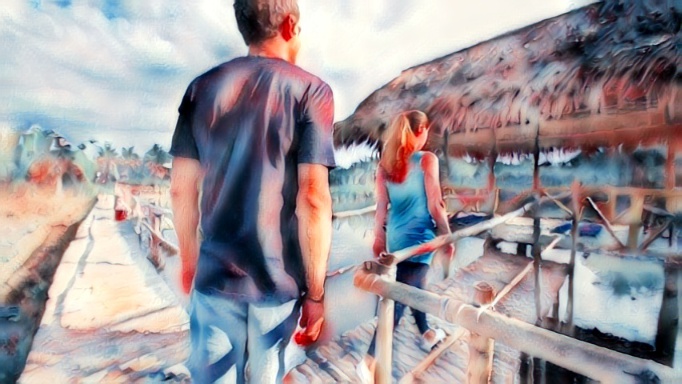}
	    &
	    \includegraphics[width=0.24\textwidth]{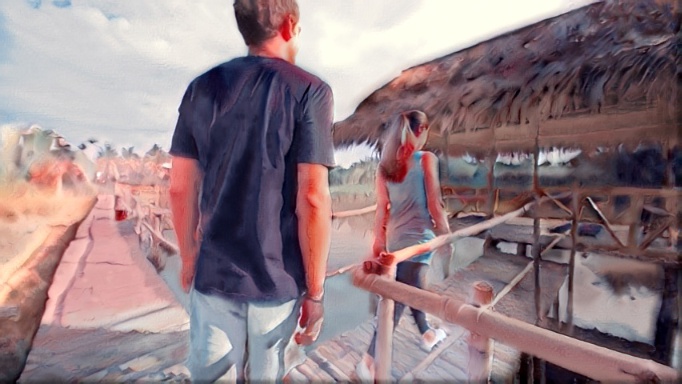}
	    &
	    \includegraphics[width=0.24\textwidth]{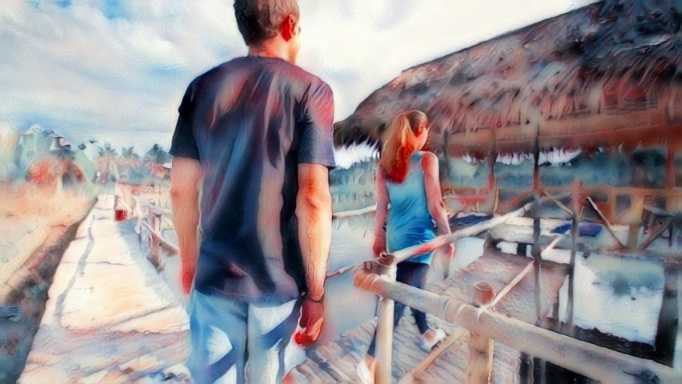}
	    \\
	    \includegraphics[width=0.24\textwidth]{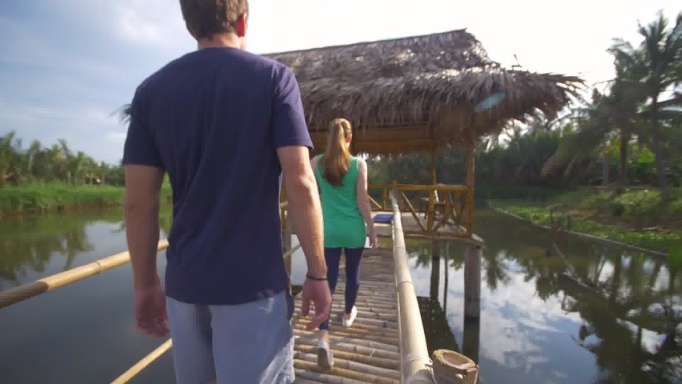}
	    &
	    \includegraphics[width=0.24\textwidth]{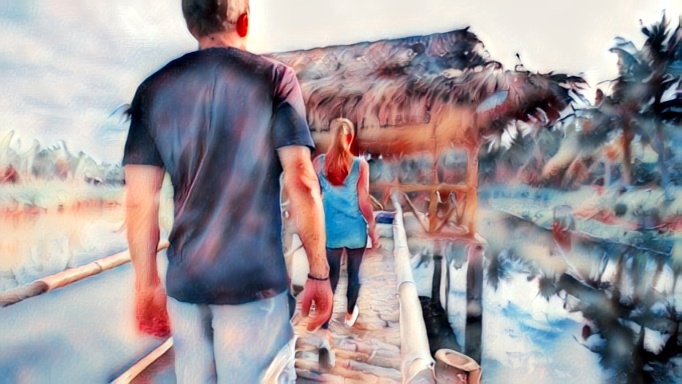}
	    &
	    \includegraphics[width=0.24\textwidth]{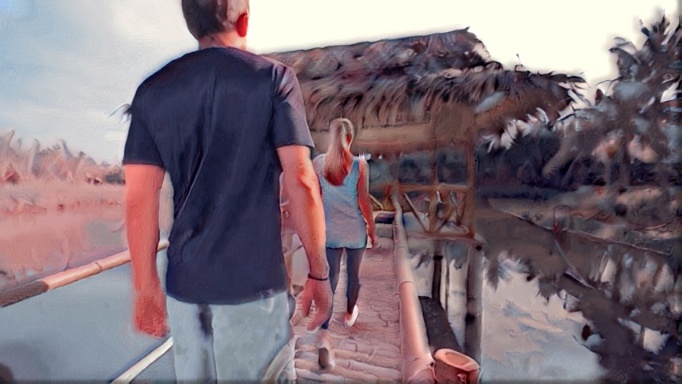}
	    &
	    \includegraphics[width=0.24\textwidth]{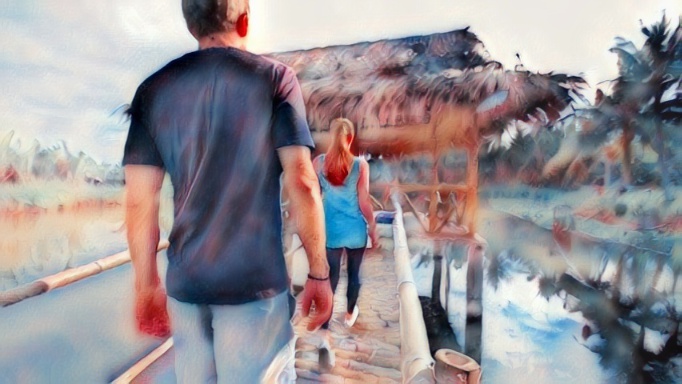}
	    \\
	    (a) Original frames &
	    (b) Processed frames & 
	    (c) Bonneel~\etal~\cite{Bonneel-TOG-2015} &
	    (d) Ours
	\end{tabular}
	\caption{
		\textbf{Visual comparisons on style transfer.}
		We compare the proposed method with Bonneel~\etal~\cite{Bonneel-TOG-2015} on smoothing the results of WCT~\cite{WCT}.
		Our approach maintains the stylized effect of processed video and reduce the temporal flickering.
	}
	\label{fig:compare_style}
\end{figure}

\begin{figure}[t]
	\centering
	\scriptsize
	\begin{tabular}{cccc}
	    \includegraphics[width=0.24\textwidth]{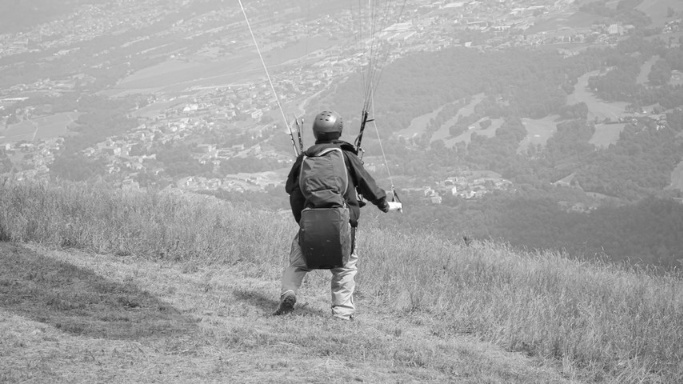}
	    &
	    \includegraphics[width=0.24\textwidth]{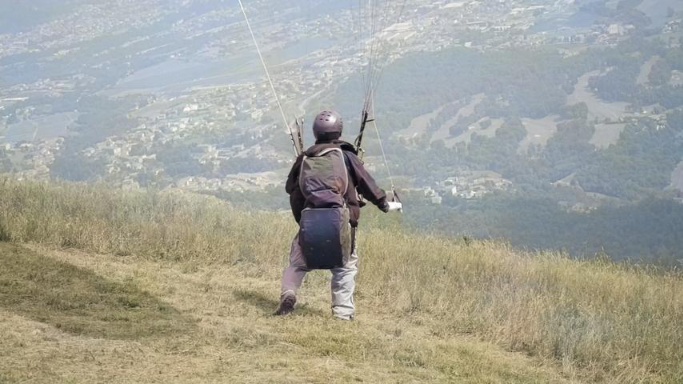}
	    &
	    \includegraphics[width=0.24\textwidth]{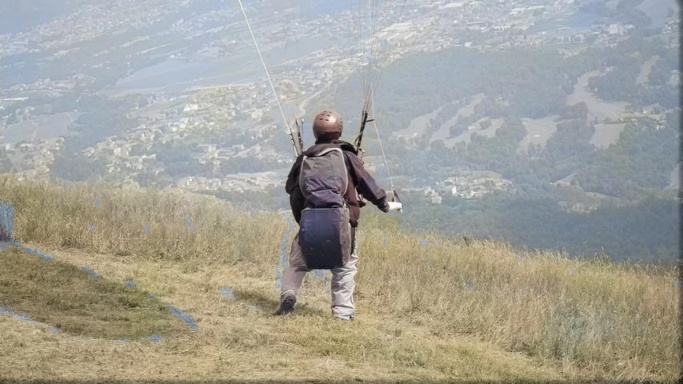}
	    &
	    \includegraphics[width=0.24\textwidth]{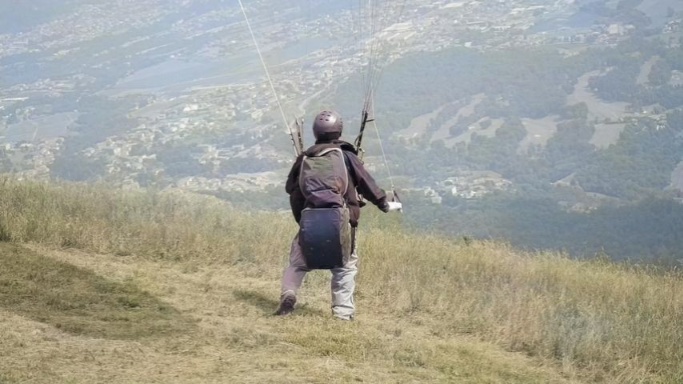}
	    \\
	    \includegraphics[width=0.24\textwidth]{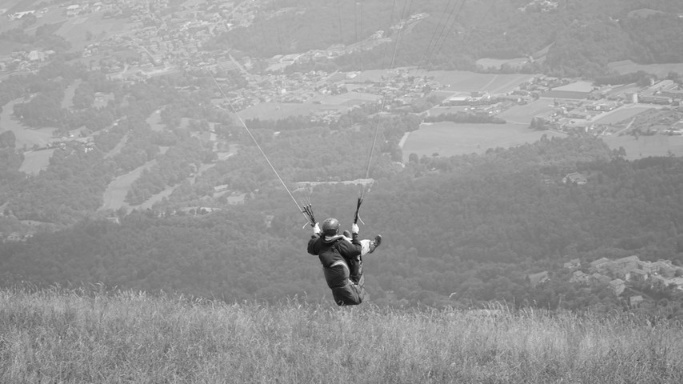}
	    &
	    \includegraphics[width=0.24\textwidth]{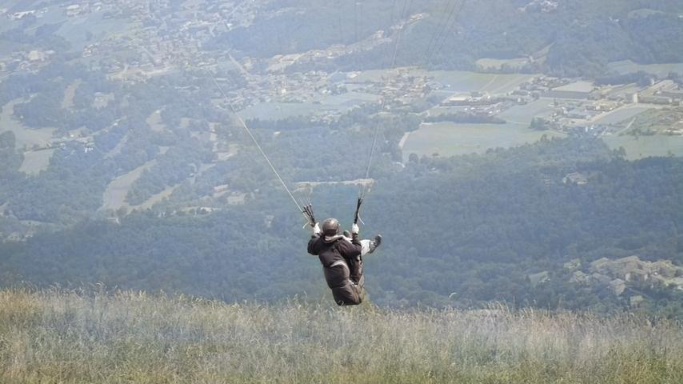}
	    &
	    \includegraphics[width=0.24\textwidth]{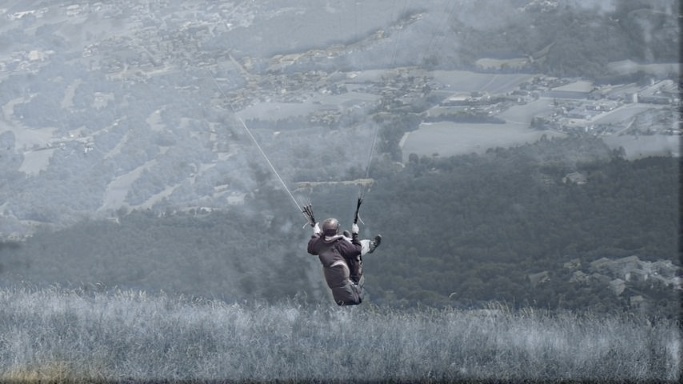}
	    &
	    \includegraphics[width=0.24\textwidth]{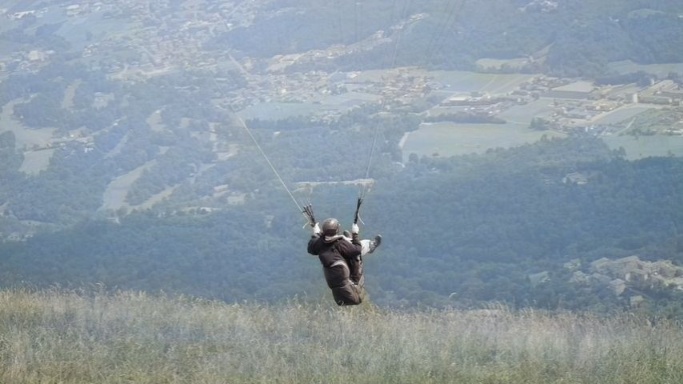}
	    \\
	    (a) Original frames &
	    (b) Processed frames & 
	    (c) Bonneel~\etal~\cite{Bonneel-TOG-2015} &
	    (d) Ours
	\end{tabular}
	\caption{
		\textbf{Visual comparisons on colorization.}
		We compare the proposed method with Bonneel~\etal~\cite{Bonneel-TOG-2015} on smoothing the results of image colorization~\cite{Iizuka-TOG-2016}.
		The method of Bonneel~\etal~\cite{Bonneel-TOG-2015} cannot preserve the colorized effect when occlusion occurs.
	}
	\label{fig:compare_colorization}
\end{figure}

\begin{figure}[t]
	\centering
	\scriptsize
	\begin{tabular}{cc}
	    \includegraphics[height=0.3\textwidth]{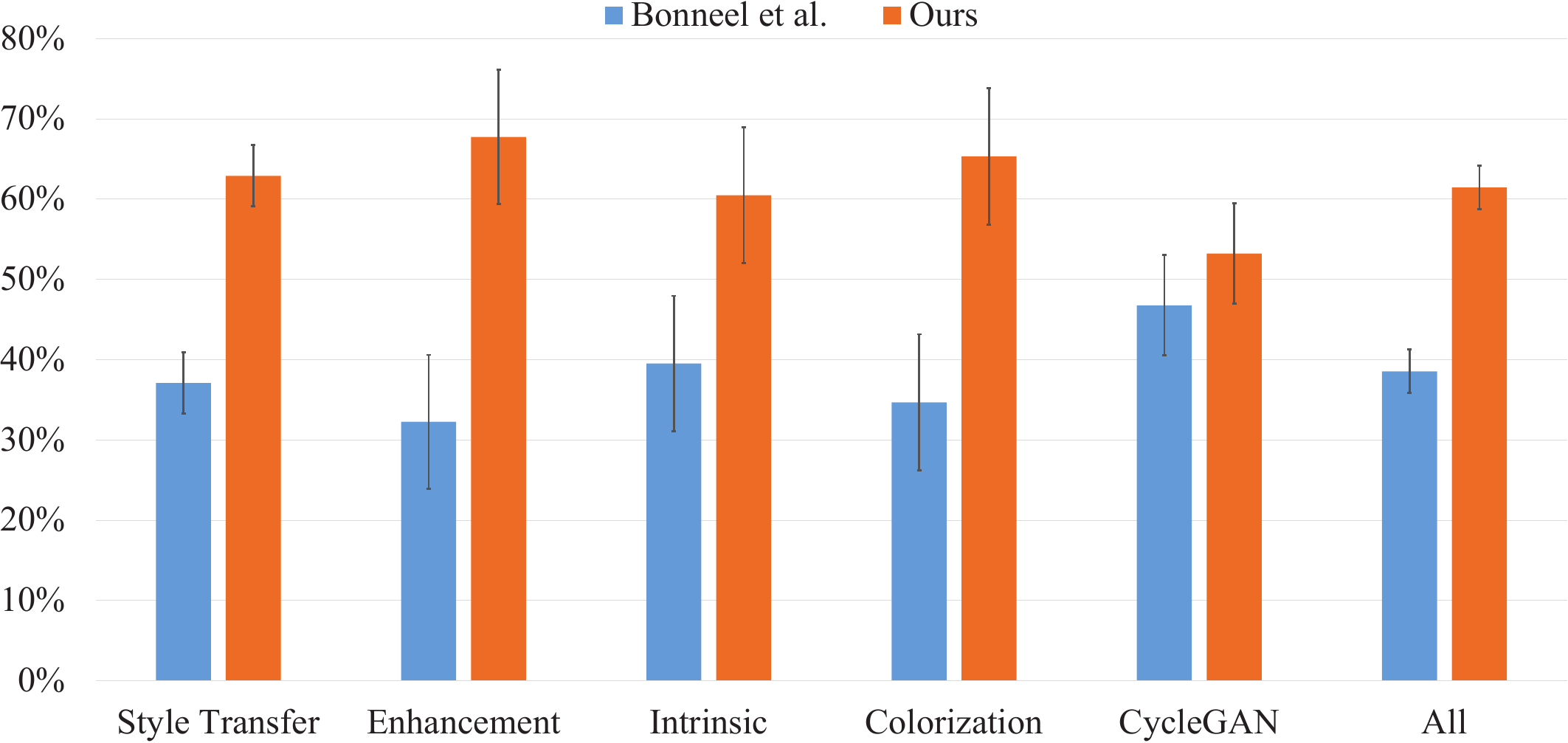}
	    &
	    \includegraphics[height=0.3\textwidth]{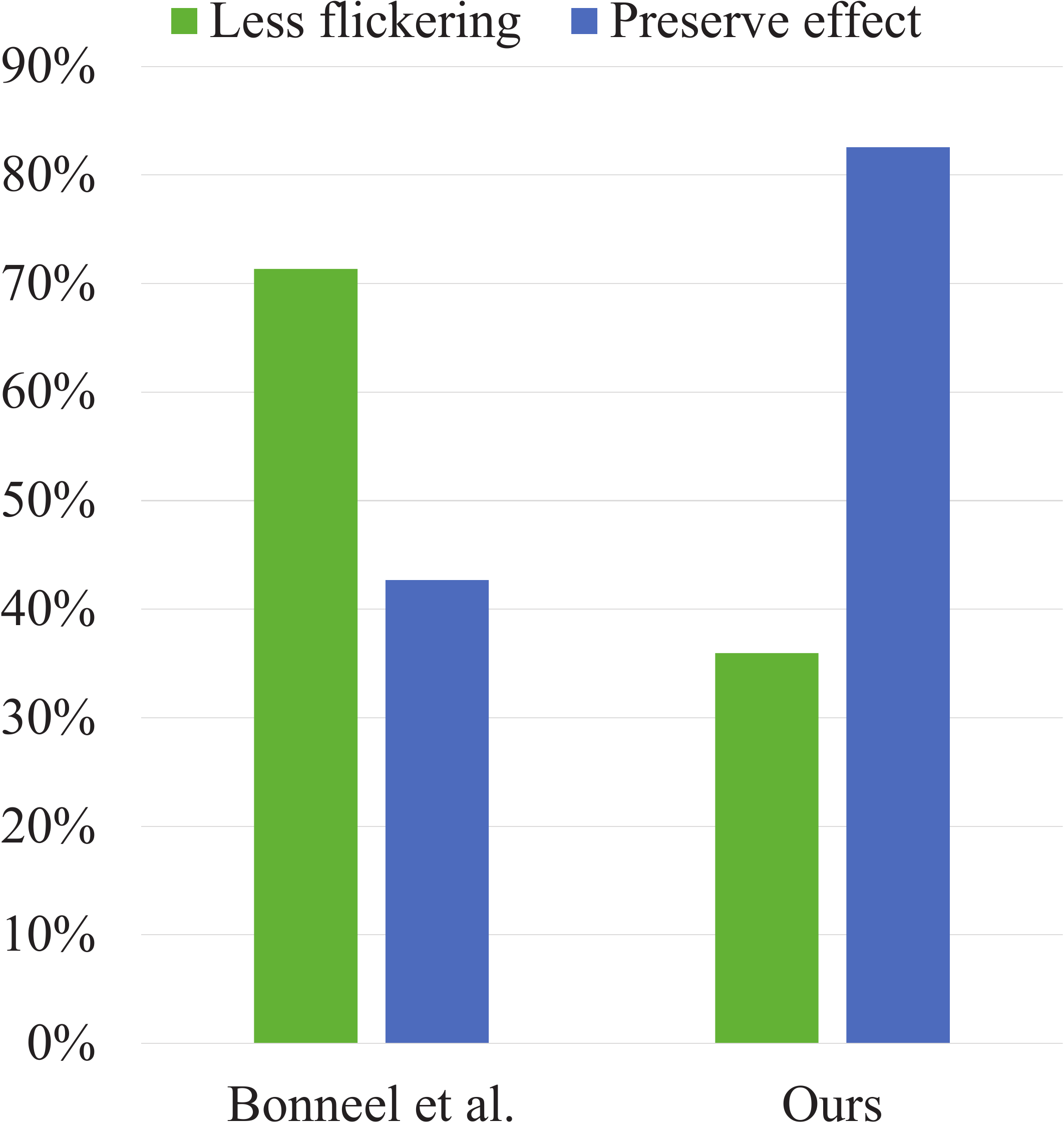}
	    \\
	    (a) Obtained votes &
	    (b) Selected reasons
	\end{tabular}
	\caption{
		\textbf{Subjective evaluation.}
		On average, our method is preferred by $62\%$ users.
		The error bars show the $95\%$ confidence interval.
	}
	\label{fig:userstudy}
\end{figure}

\subsection{Comparison with state-of-the-art methods}
\label{sec:compare_stoa}
We evaluate the temporal warping error~\eqnref{temporal_warp_error} and perceptual distance~\eqnref{perceptual_distance} on the two video test sets.
We compare the proposed method with Bonneel~\etal~\cite{Bonneel-TOG-2015} on 16 applications: 2 styles of Johnson~\etal~\cite{Johnson-ECCV-2016}, 6 styles of WCT~\cite{WCT}, 2 enhancement models of Gharbi~\etal~\cite{DBL}, reflectance and shading layers of Bell~\etal~\cite{Bell-TOG-2014}, 2 photo-to-painting models of CycleGAN~\cite{CycleGAN} and 2 colorization algorithms~\cite{Iizuka-TOG-2016,Zhang-ECCV-2016}.
We provide the average temporal warping error and perceptual distance in~\tabref{warp_error} and~\tabref{perceptual}, respectively.
In general, our results achieves lower perceptual distance while maintains comparable temporal warping error with the results of Bonneel~\etal~\cite{Bonneel-TOG-2015}.

We show visual comparisons with Bonneel~\etal~\cite{Bonneel-TOG-2015} in~\figref{compare_style} and \ref{fig:compare_colorization}.
Although the method of Bonneel~\etal~\cite{Bonneel-TOG-2015} produces temporally stable results, the assumption of identical gradients in the processed and original video leads to overly smoothed contents, for example from stylization effects.
Furthermore, when the occlusion occurs in a large region, their method fails due to the lack of a long-term temporal constraint.
In contrast, the proposed method dramatically reduces the temporal flickering while maintaining the perceptual similarity with the processed videos.
We note that our approach is not limited to the above applications but can also be applied to tasks such as automatic white balancing~\cite{Hsu-TOG-2008}, image harmonization~\cite{Bonneel-2015} and image dehazing~\cite{He-PAMI-2011}.
Due to the space limit, we provide more results and videos on our project website.

\subsection{Subjective evaluation}

We conduct a user study to measure user preference on the quality of videos.
We adopt the pairwise comparison, i.e., we ask participants to choose from a pair of videos.
In each test, we provide the original and processed videos as references and show two results (Bonneel~\etal~\cite{Bonneel-TOG-2015} and ours) for comparisons.
We randomize the presenting order of the result videos in each test.
In addition, we ask participants to provide the reasons that they prefer the selected video from the following options: (1) The video is less flickering. (2) The video preserves the effect of the processed video well.

We evaluate all 50 test videos with the 10 test applications that were held out during training.
We ask each user to compare 20 video pairs and obtain results from a total of 60 subjects.
\figref{userstudy}(a) shows the percentage of obtained votes, where our approach is preferred on all 5 applications.
In~\figref{userstudy}(b), we show the reasons when a method is selected.
The results of Bonneel~\etal~\cite{Bonneel-TOG-2015} are selected due to temporal stability, while users prefer our results as we preserve the effect of the processed video well.
The observation in the user study basically follows the quantitative evaluation in~\secref{compare_stoa}.

\subsection{Execution time}
We evaluate the execution time of the proposed method and Bonneel~\etal~\cite{Bonneel-TOG-2015} on a machine with a 3.4 GHz Intel i7 CPU (64G RAM) and an Nvidia Titan X GPU.
As the proposed method does not require computing optical flow at test time, the execution speed achieves 418 FPS on GPU for videos with a resolution of $1280 \times 720$.
In contrast, the speed of Bonneel~\etal~\cite{Bonneel-TOG-2015} is 0.25 FPS on CPU.

\subsection{Limitations and discussion}
Our approach is not able to handle applications that generate entirely different image content on each frame, e.g., image completion~\cite{ContextEncoder} or synthesis~\cite{Chen-ICCV-2017b}.
Extending those methods to videos would require incorporating strong video priors or temporal constraints, most likely into the design of the specific algorithms themselves. 

In addition, in the way the task is formulated there is always a trade-off between being temporally coherent or perceptually similar to the processed video. 
Depending on the specific effect applied, there will be cases where flicker (temporal instability) is preferable to blur, and vice versa.
In our current method, the user can choose a model based on their preference for flicker or blur, but an interesting area for future work would be to investigate perceptual models for what is considered acceptable flicker and acceptable blur.
Nonetheless, we use the same trained model (same parameters) for all our results and showed clear viewer preference over prior methods for blind temporal stability.

\section{Conclusions}

In this work, we propose a deep recurrent neural network to reduce the temporal flickering problem in per-frame processed videos.
We optimize both short-term and long-term temporal loss as well as a perceptual loss to reduce temporal instability while preserving the perceptual similarity to the processed videos.
Our approach is agnostic to the underlying image-based algorithms applied to the video and generalize to a wide range of unseen applications.
We demonstrate that the proposed algorithm performs favorably against existing blind temporal consistency method on a diverse set of applications and various types of videos.

\section*{Acknowledgements}
This work is supported in part by the NSF CAREER Grant $\#$1149783, NSF Grant No. $\#$ 1755785, and gifts from Adobe and Nvidia.

\clearpage

\bibliographystyle{splncs04}
\bibliography{reference}
\end{document}